\documentclass{article}
\usepackage{spconf,amsmath,graphicx}
\usepackage[colorlinks,linkcolor=blue]{hyperref}
\usepackage{caption}
\usepackage{multirow}
\usepackage{float}

\usepackage{amsmath}
\usepackage[psamsfonts]{amssymb}
\title{Spatial-temporal Transformers for EEG Emotion Recognition}
\name{Jiyao Liu, Hao Wu, Li Zhang, Yanxi Zhao}
\address{School of Computer Science, \\
 Northwestern Polytechnical University, Xi'an, China }
 
\begin{document}
 
\maketitle
 
\begin{abstract}
Electroencephalography (EEG) is a popular and effective tool for emotion recognition. However, the propagation mechanisms of EEG in the human brain and its intrinsic correlation with emotions are still obscure to researchers. This work proposes four variant transformer frameworks~(spatial attention, temporal attention, sequential spatial-temporal attention and simultaneous spatial-temporal attention) for EEG emotion recognition to explore the relationship between emotion and spatial-temporal EEG features. Specifically, spatial attention and temporal attention are to learn the topological structure information and time-varying EEG characteristics for emotion recognition respectively. Sequential spatial-temporal attention does the spatial attention within a one-second segment and temporal attention within one sample sequentially to explore the influence degree of emotional stimulation on EEG signals of diverse EEG electrodes in the same temporal segment. The simultaneous spatial-temporal attention, whose spatial and temporal attention are performed simultaneously, is used to model the relationship between different spatial features in different time segments. The experimental results demonstrate that simultaneous spatial-temporal attention leads to the best emotion recognition accuracy among the design choices, indicating modeling the correlation of spatial and temporal features of EEG signals is significant to emotion recognition.
\end{abstract}
\section{Introduction}
\label{sec:intro}
EEG emotion recognition is to detect the current emotional states of the subjects \cite{bos2006eeg},~\cite{cowie2001emotion},~\cite{gunes2011emotion}. In recent years, with the development of deep learning and the availability of EEG data, many emotion recognition methods based on neural networks have dominated the state-of-art position \cite{al2017classification,li2018hierarchical}.  
In general, the EEG signals collected by a spherical EEG cap have three-dimensional characteristics which are spatial, spectral and temporal respectively. Many researchers have drawn attention to how to effectively utilize time-varying spatial and temporal features from multi-channel brain signals. 

In order to model the space relationships among multi-channel EEG signals,
a hierarchical convolutional neural network (CNN)~ is proposed by Li et al.~\cite{li2018hierarchical} to capture spatial information among different channels. A deep CNN model is present by Zhang et al.~\cite{zhang2017eeg} to capture the spatio-temporal robust feature representation of the raw EEG data stream for motion intention classification. A utilization of multi-layer CNN with no full connection layers is proposed by Lawhern et al.~\cite{lawhern2018eegnet} for P300-based oddball recognition task, finger motor task and motor imagination task. 

Considering  the change of EEG signals over time, an Echo State Network (ESN) is present by Fourati et al.~\cite{fourati2017optimized}, ESN used recursive layer 
to map the EEG signal into the high-dimension state space. A two-layer long short term memory (LSTM), which uses EEG signal as the input, is adopted by Alhagry et al.~\cite{alhagry2017emotion} and obtain promising EEG emotion classification results. A deep recursive convolutional neural network (R-CNN) is present by Bashivan et al.~\cite{bashivan2015learning}, the proposed R-CNN gets a satisfactory result on the task mental load classification based on EEG signal. 

Most of the above works are on the basis of convolution or recursive operation. CNN is good at modeling local receptive field message, while pays less attention to the global information. Recurrent Neural Networks (RNN) network is relatively weak to capture the spatial information and its parallel computational efficiency is slower. To solve the above weaknesses, some works lead 
attention mechanism into CNN and RNN.

Since different spatial-temporal features have different contributions to emotion recognition, they should be assigned to different weight in the classifier recognizing emotions. 
A LSTM with attention mechanism is proposed by Kim et al.~\cite{kim2020eeg}, the network assigns weights to the emotional states appearing at specific moments to conduct two-level and three-level classification on the valence and arousal emotion models. A fresh multi-channel model on the basis of sparse graphic attention long short term memory (SGA-LSTM) is present by  Liu et al.~\cite{liu2019sparse} to classify EEG emotion. 

As is mentioned above, existing works have attained gratifying results. However,The transmission characteristic as well as spatial-temporal relevance of different EEG electrodes are more or less neglected in most of them. The changeless size kernels in convolution operation~\cite{krizhevsky2012imagenet} may damage the spatial correlation of EEG signals. Though the RNN operation~\cite{rumelhart1986learning} takes the temporal features of EEG signals into consideration, it ignores the spatial relation among EEG electrodes. Furthermore, on account of the diverse impedance of various brain areas, there may be a slight error in time between the EEG signal displayed by the EEG collection device and the real EEG signal, that is, EEG signals may delay varies with different EEG electrodes.

To deal with the mentioned issues, we present a fresh EEG emotion transformer~(EeT) framework built exclusively on self-attention blocks. The variants of self-attention block include spatial~(S) attention, temporal~(T) attention, sequential spatial-temporal~(S-T) attention and simultaneous spatial-temporal~(S+T) attention. The spatial attention is to learn the spatial structure information. The temporal attention is to learn the correlation between EEG signals and emotional stimuli as well as temporal changes. The sequential spatial-temporal attention is to do spatial attention within the same time segment and temporal attention among different time segments in one sample. The simultaneous spatial-temporal attention is to do the two attention simultaneously. Experimental setups are elaborately picked to study the effects of spatial and temporal EEG signals on emotion recognition, and whether there is some correlation between the features of different channels at different time segments. 

\section{Variants of EEG Emotion Transformers}
\subsection{Framework of EeT}
\begin{figure}[ht]
\captionsetup{font={scriptsize}} 
\centering
\includegraphics [scale=0.39]{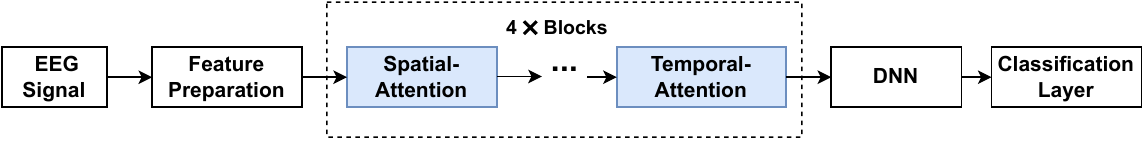}
\centering
\caption{Overview of EEG Emotion Recognition Transformer~(EeT) }
\label{fig: overview}
\centering
\end{figure}

EEG-based emotion recognition is to classify the emotion states according to the EEG signal. As illustrated in Fig.~\ref{fig: overview}, the overview of EeT includes four modules, namely the feature preparation module, the spatial-temporal attention module, deep neural network~(DNN) module and classification module. We focus on the design of self-attention module which includes spatial attention, temporal attention, sequential spatial-temporal attention and simultaneous spatial-temporal attention. 

Emotion recognition based on EEG is to learn a function $f$ which maps the raw signals to emotion tags:

\begin{equation}
Y=f(X^{'})
\label{DKL}
\end{equation}
where $X^{'}$ represents the EEG features. $f$ represents the mapping function i.e., convolutional neural network transformation. $Y \in \{y_1, y_2, ... , y_n\}$ represents the emotional tags. In our work, the cross entropy is adopted as the loss function, which can be defined as:
\begin{equation}
L=-\sum_{c=1}^{C}y_{c}log(y^{'}_{c})
\label{DKL}
\end{equation}
where $L$ denotes the loss function of the EEG emotion classification, $C$ denotes the number of emotion states, $y_{c}$ is the ground-truth emotion tag and $y^{'}_{c}$ is the predictor of neural networks. 

\subsection{Preprocessing}
The EEG features can be denoted as $X=(F_{1},F_{2},...,F_{T}) \in \mathbb{R}^{C\times S \times T}$, where $C$ is 5, equals to the number of frequency bands~($\delta$[1-4Hz], $\theta$ [4-8Hz], $\alpha$  [8-14Hz], $\beta$ [14-31Hz], and $\gamma$ [31-50 Hz]) used to compute the EEG features. $S$ equals to  the number of electrodes in the EEG cap and $T$ equals to the number of the time slots in a EEG sample. $F_{t}=(B_{1},B_{2},...,B_{S})\in \mathbb{R}^{C\times S} (t\in \{1,2,...,T\})$ is the one-second EEG feature. $B_{s}=(b_{1},b_{2},...,b_{C})\in \mathbb{R}^{C} (s\in \{1,2,...,S\})$ presents the feature of one EEG channel. Specifically, we map the $S$ electrodes into a  $V \times H$ matrix according to the layout of the EEG cap in order to preserve the spatial topology structure information of the EEG cap, then we take advantage of the linear interpolation~\cite{schlitt1995evaluation} to replenish the spatial information which are not collected by EEG acquisition equipment. The re-organized feature can be represent as $F^{'}_{t}=(B_{1},B_{2},...,B_{V \times H})\in \mathbb{R}^{C\times V \times H}~(t\in {1,2,...,T})$ for a one-second EEG slice and $X^{'}=(F_{1},F_{2},...,F_{T}) \in \mathbb{R}^{C\times V \times H \times T}$ for the whole EEG sample.


\subsection{Positional Encoding for Spatial EEG Electrodes}

We divided the EEG feature of each second $F_{t} (i=1,2,3..S)$ into $G$ non-overlapping regions,
just like the different brain regions in neuroscience.
Here we regroup the $V \times H$ matrices into region sequences,  the size of each divided region is $P \times P$, so we get $G=VH/P^{2}$ regions. Each region is flatten into a vector $I(x)_{(p,t)}\in \mathbb{R}^{5P^{2}}$ with  $p=1,2...,G$ representing spatial layout of EEG electrodes and $t=1,2,...T$ denoting the index over seconds. Then we linearly map each region $I(x)_{(p,t)}$
into a latent vector $z_{(p,t)}^{(0)}\in \mathbb{R}^{D}$ by means of learnable matrix $M\in\mathbb{R}^{D\times 5P^{2}}$:
\begin{equation}
z_{(p,t)}^{(0)}={M \otimes I(x)}_{(p,t)}+e_{(p,t)}^{position}
\label{DKL}
\end{equation}

\noindent where $\otimes$ is matrix multiplication and $e_{(p,t)}^{pos}\in\mathbb{R}^{D}$ stands for a learnable position embedding to encode the spatial-temporal position of each brain region. The resulting sequence of embedding vectors $z_{(p,t)}^{(0)}$ stands for the input to
the next layer of the self-attention block. Note that $z^i$ is output of the $ith$ layer in self-attention block. $p=1,...G$ and $t=1,...,T$ are the spatial locations and indexes over time segments respectively.
 \subsection{Query-Key-Value Mechanism}
Our Transformer consists of $L$ encoding blocks. Instead of performing a single attention function,  we use different projected versions of
queries, keys and values to perform the attention function in parallel, which is called  multi-head attention. At each block $l$, the query/key/value vectors are computed for each region from the representation $z_{(p,t)}^{(l-1)}$ encoded by the preceding block:
\begin{equation}
q_{(p,t)}^{(l,a)}=W_{Q}^{(l,a)}z_{(p,t)}^{(l-1)}\in\mathbb{R}^{D_{h}}
\end{equation}
\begin{equation}
k_{(p,t)}^{(l,a)}=W_{K}^{(l,a)}z_{(p,t)}^{(l-1)}\in\mathbb{R}^{D_{h}}
\end{equation}
\begin{equation}
v_{(p,t)}^{(l,a)}=W_{V}^{(l,a)}z_{(p,t)}^{(l-1)}\in\mathbb{R}^{D_{h}}
\end{equation}
$z_{(p,t)}^{(l-1)}$, which is the output of the previous block, need to be layer normalized before the above operations. $a=1,2...,A$ denotes an index over multiple attention heads and $A$ is the number of attention heads.

\subsection{Variants of Attention Mask Learning }
The variants of self-attention block include spatial attention (S), temporal attention (T), sequential spatial-temporal attention (S-T) and simultaneous spatial-temporal attention (S+T). The spatial  attention is to learn the spatial structure information while the temporal attention is to the relationship between EEG and time. The sequential spatial-temporal attention is the concatenation of two operations. The simultaneous spatial-temporal attention is to do the two operations simultaneously. 
\subsubsection{Spatial Attention}
 
 In the case of spatial attention, the self-attention weights $\alpha_{(p,t)}^{(a,l)} \in \mathbb{R}^{N+1} $ for query brain region $(p,t)$ are given by:

\begin{equation}
{\alpha}_{(p, t)}^{(l, a) \text { spatial }}=\sigma\left(\frac{q_{(p, t)}^{(l, a)^{\top}}}{\sqrt{D_{h}}} \cdot\left[k_{(0,0)}^{(l, a)}\left\{k_{\left(p^{\prime}, t\right)}^{(l, a)}\right\}_{p^{\prime}}\right]\right)
\label{7}
\end{equation}

where ${p}'$ denotes the index of the brain regions. $\sigma$ denotes the softmax activation function. The formula is to consider that different brain regions react differently under the same emotional stimulation thus different weights are given to the features of different brain regions.
 \subsubsection{Temporal Attention}
 For the temporal attention, the self-attention weights $\alpha_{(p,t)}^{(l,a)} \in \mathbb{R}^{T+1} $ for query brain region $(p,t)$ are given by:
\begin{equation}
{\alpha}_{(p, t)}^{(l, a) \text { temporal }}=\sigma\left(\frac{q_{(p, t)}^{(l, a)^{\top}}}{\sqrt{D_{h}}} \cdot\left[k_{(0,0)}^{(l, a)}\left\{k_{\left(p, t^{\prime}\right)}^{(l, a)}\right\}_{t^{\prime}}\right]\right)
\label{8}
\end{equation}
where ${t}'$ denotes the index of the time slots. In this formula, different weights are given to the of different time slots in consideration of the change of EEG signals with emotional stimulation and time.
 
\subsubsection{Sequential Spatial-Temporal~(S-T) Attention}
The sequential spatial-temporal attention is to do spatial attention within the same time segment and temporal attention among different time segments. Firstly, the spatial self-attention weights are calculated as Eq.~\ref{7}. Then the the temporal attention weights are learned by Eq.~\ref{8} from the output of spatial attention layer. (S-T) Attention comprehensively considers the attention of space and time, but the default is that the spatial features in the same time period are closely related, while the features of different brain in different time are weakly related.
 
\subsubsection{Simultaneous Spatial-Temporal~(S+T) Attention}
The simultaneous spatial-temporal~(S+T) attention is to do spatial and temporal attention simultaneously. The self-attention weights $\alpha_{(p,t)}^{(a,l)} \in \mathbb{R}^{NT+1} $ for query brain region $(p,t)$ are given by:
\begin{equation}
{\alpha_{(p,t)}^{(l,a)}}=\sigma \left (  \frac{{q_{(p,t)}^{(l,a)}}^T}{\sqrt D_{h}} \right )  \cdot \left [ k_{(0,0)}^{(l,a)}\left \{   k_{({p}', {t}')}^{(l,a)}\right \}_{\begin{matrix}
{p}'=1,...,N \\ {t}'=1,...,T
\end{matrix}}\right ]
\label{DKL}
\end{equation}
Different from S-T Attention, which regards space and time separately, the S+T attention considers that the spatial information in the same time point and different time points are both strongly correlated.

\subsection{Multi-head Attention Recalibration} The encoding $z_{(p,t)}^{(l)}$
at block $l$ is obtained by
the first computing the weighted sum of value vectors using
self-attention coefficients from each attention head:
\begin{equation}
    {s_{(p,t)}^{(l,a)}}= \alpha _{(p,t),(0,0)}^{(l,a)} v_{(0,0)}^{(l,a)} +  \sum_{{p}'=1}^{N} \sum_{{t}'=1}^{F}\alpha _{(p,t),({p}',{t}')}^{(l,a)}v_{({p}',{t}')}^{(l,a)},
\end{equation}
As is mentioned above, $a$ denotes an index over multiple attention heads and $l$ is the index of the blocks.
Then, the concatenation of these vectors from all heads is projected and passed through an MLP, using residual connections after each operation:
\begin{equation}
 {{z}'}_{(p,t)}^{l} = {W_{O}^{(l-1)}} \begin{bmatrix}
s_{(p,t)}^{(l,1)}\\ . 
\\ .
\\ .
\\ s_{(p,t)}^{(l,A )}

\end{bmatrix}  + z_{(p,t)}^{(l-1)},
\end{equation}
where $W_{O}$ is the \textbf{Value} of $z_{(p,t)}^{(l-1)}$ by concatenating $v_{(p,t)}^{(l,a)}$.
\begin{equation}
  z_{(p,t)}^{l} = MLP({{z}'}_{(p,t)}^{l}) + {{{z}'}_{(p,t)}}^{l}  
\end{equation}
The ${{z}'}_{(p,t)}^{l}$ goes through the MLP layer to get the output of $lth$ layer. 
\section{Experiments}
\subsection{Datasets}
We validate our model on SEED \cite{zheng2015investigating, duan2013differential}, SEED-IV \cite{zheng2018emotionmeter} and Deap \cite{koelstra2011deap} databases.

\textbf{Deap} dataset is a open source dataset including diverse physiological signals with emotion evaluations provided by the research team of Queen Mary University in London. It records the EEG, ECG, EMG and other bioelectrical signals of 32 subjects induced by watching 40 one-minute music videos of different emotional tendencies. The subjects evaluated the videos’ emotion categories on scale of one to nine in dimension of arousal, valence, liking, dominance and familiarity. Valence reports the degree
of subjects' joy, the greater the valence value, the higher the joy degree. Arousal reports the subjects' emotional intensity, the higher the arousal value, the more intense and perceptible the emotion. The rating value from small to large indicates the emotion metric is from negative
to positive or from weak to strong. The 40 stimulus videos include 20 high valence/arousal stimuli and 20 low valence/arousal stimuli.

\textbf{SEED} contains three different categories of emotion, namely positive, negative, and neutral. Fifteen participants’ EEG data of the dataset were collected while they were watching the stimulus videos. The videos are carefully selected and can elicit a single desired target emotion. With an interval of about one week, each subject participated in three experiments, and each session contained 15 film clips.
The participants are asked to give feedback immediately after each experiment. The EEG signals of 62 channels are recorded at a sampling frequency of 1000 Hz and down-sampled with 200 Hz. The Differential entropy~\cite {duan2013differential} DE features are pre-computed over different frequency bands for each sample in each channel.

\textbf{SEED-IV} contains four different categories of emotions, including happy, sad, fear, and neutral emotion. The experiment consists of 15 participants. Three experiments are designed for each participant on different days, and each session contains 24 video clips and six clips for each type of emotion in each session.  After each experiment, the subjects are asked to give feedback, while 62 EEG signals of the subjects are recorded. The EEG signals are sliced into 4-second non-overlapping segments and down-sampled with 128 Hz. The DE feature is also pre-computed over five frequency bands in each channel.
\subsection{Experimental Setup}
We train our model on NVIDIA RTX 2080 GPU. Cross entropy loss is used as the loss function. The optimizer is Adam. The initial learning rate is set to 1e-3 with multi-step decay to 1e-7. The number of the attention blocks is set to 4 and the length of each sample is set to 10s. We conduct experiments on each subject. For each experiment, we randomly shuffle the samples and use 5-fold cross validation. The ratio of the training set to test set is 9:6.

\subsection{Compared Models}
We compare the proposed EeT with the following competitive models.

SVM~\cite{suykens1999least} is a least squares support vector machine classifier. 
DBN~\cite{zheng2014eeg} is deep Belief Networks investigate the critical frequency bands and channels. DGCNN~\cite{song2018eeg} is Dynamical Graph Convolutional Neural Networks model the multichannel EEG features.
BiDANN~\cite{li2018novel} is bi-hemispheres domain adversarial neural network maps the EEG feature of both hemispheres into discriminative feature spaces separately.
BiHDM~\cite{li2020novel} is bi-hemispheric discrepancy model learns the asymmetric differences between two hemispheres for EEG
emotion recognition.
3D-CNN with PST-Attention~\cite{liu2021positional} is a self-attention module combined with 3D-CNN to learn critical information among different dimensions of EEG feature. LSTM~\cite{alhagry2017emotion} is a time series model for identifying continuous dimension emotion. 3D-CNN~ \cite{salama2018eeg} is 3D-CNN model to recognize arousal and valence. BT~\cite{chen2019accurate} is deep convolution neural network for continuous dimension emotion recognition.

\subsection{Experimental Results and Analysis}

\begin{table}[th]
\tiny
\centering
\captionsetup{font={scriptsize}} 
\caption{Experimental Results on DEAP Dataset}
\resizebox{\linewidth}{!}{
\begin{tabular}{cclcl}
\hline
\multirow{2}{*}{\textbf{Models}} & \multicolumn{2}{c}{Arousal} & \multicolumn{2}{c}{Valence} \\
                                 & acc(\%)          & F1       & acc(\%)          & F1       \\ \hline
LSTM\cite{alhagry2017emotion}                             & 85.65            & -        & 85.45            & -        \\
3D-CNN \cite{salama2018eeg}                          & 88.49            & -        & 87.44            & -        \\
BT\cite{chen2019accurate}                              & 86.18            & -        & 86.31            & -        \\
EeT$\sim$(S+T Attention)         & \textbf{93.34}   & 0.9326   & \textbf{92.86}   & 0.9196   \\ \hline      
\end{tabular}
}
\label{results12}
\end{table}
Table~\ref{results12} presents the average accuracy (acc) and F1 value
(F1) of the compared models for EEG based emotion recognition on the DEAP datasets. Compared with  3D-CNN~\cite{salama2018eeg}, the acc of the proposed simultaneous spatial-temporal ~(S+T) attention EeT framework have 4.85\%/5.42\% improvement.  EeT with S+T attention also gets superior performance compared with other competitive models.
\begin{table}[th]
\huge
\centering
\captionsetup{font={scriptsize}} 
\caption{Experimental Results on SEED and SEED-IV Dataset}
\resizebox{\linewidth}{!}{
\begin{tabular}{cclcl}
\hline
\multirow{2}{*}{\textbf{Models}} & \multicolumn{2}{c}{\textbf{SEED}} & \multicolumn{2}{c}{\textbf{SEED-IV}} \\
                                 & Mean (\%)         & Std (\%)        & Mean (\%)          & Std (\%)          \\ \hline
SVM \cite{suykens1999least}                     & 83.99            & 9.72           & 56.61             & 20.05            \\
DBN \cite{zheng2014eeg}                     & 86.08            & 8.34           & 66.77             & \textbf{7.38}             \\
DGCNN \cite{song2018eeg}                   & 90.40            & 8.49           & 69.88             & 16.29            \\
BiDANN \cite{li2018novel}                  & 92.38            & 7.04           & 70.29             & 12.63            \\
BiHDM \cite{li2020novel}                   & 93.12            & 6.06           & 74.35             & 14.09            \\
3D-CNN with PST-Attention\cite{liu2021positional}                        & 95.76            & 4.98           & 82.73             & 8.96             \\ 
\textbf{EeT~(S+T Attention)}                   & \textbf{96.28}            & \textbf{4.39 }          & \textbf{83.27}             & 8.37            \\ \hline
\end{tabular}
}
\label{results}
\end{table}

Table \ref{results} presents the average accuracy (Mean) and standard deviation (Std) of the compared models for EEG based emotion recognition on SEED and SEED-IV datasets. Compared with 3D-CNN with PST-Attention, the means of the proposed joint spatial+temporal~(S+T) attention EeT framework have 0.52\%/0.54\% improvements on SEED and SEED-IV. The Stds of Eet with S+T attention achieve 0.59\%/0.59\% reductions on SEED and SEED-IV respectively compared with those of 3D-CNN with PST-Attention. Moreover, EeT with S+T attention gets superior performance compared with other competitive models.

\begin{figure}[ht]
\captionsetup{font={scriptsize}} 
\centering
\includegraphics [scale=0.3]{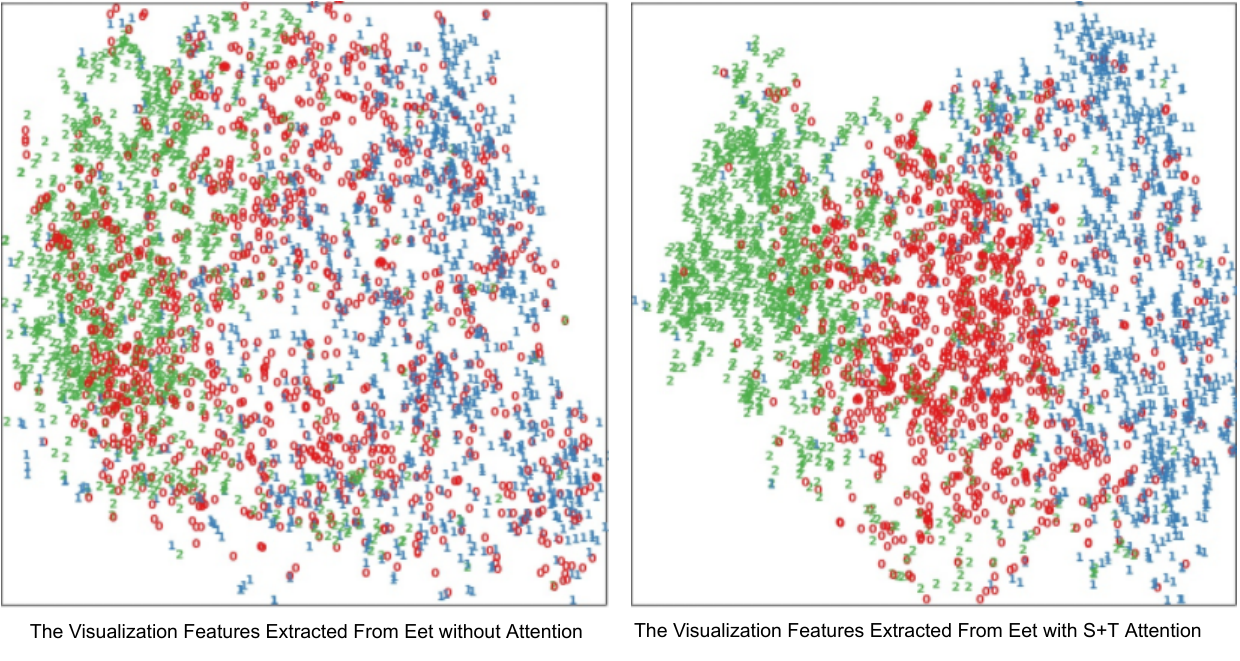}
\centering
\caption{The Visualization of High Level Features Extracted From Eet}
\label{fig: att}
\centering
\end{figure}
We use t-SNE~\cite{van2008visualizing} to visualize the high-level bottleneck features from the well trained Eet. As shown in Fig.~\ref{fig: att}, different colors represent different emotional labels, the distances of different classes in the high-level feature space of Eet with S+T Attention~(the right part) are more dispersed than that of EeT without attention~(the left part), which demonstrates that the high-level features learned with simultaneous spatial-temporal attention is more discriminative.
\subsection{Ablation Experiments}
\begin{table}[th]
\captionsetup{font={scriptsize}} 
\centering
\caption{Experimental Results of Variant Transformers}
\resizebox{\linewidth}{!}{
\begin{tabular}{cclcl}
\hline
\multirow{2}{*}{\textbf{Models}} & \multicolumn{2}{c}{\textbf{SEED}} & \multicolumn{2}{c}{\textbf{SEED-IV}} \\
                                 & Mean (\%)         & Std (\%)        & Mean (\%)          & Std (\%)          \\ \hline
S Attention                    & 93.14            & 9.31           &73.31             & 13.67            \\
T Attention                  & 92.74           & 10.21           & 72.37             & 12.83            \\
S-T Attention                      & 95.65            & 6.73           & 80.31             &  8.51          \\
\textbf{S+T Attention}                   & \textbf{96.28}            & \textbf{4.39}           & \textbf{83.27}             & \textbf{8.37}            \\
\hline
\end{tabular}
}
\label{results2}
\end{table}
Table \ref{results2} presents the results of different variants of the proposed transformer framework, from which we can see that Joint Spatial-Temporal Attention gets the best results, achieving 0.63\%/2.96\% improvements compared with the second best variant, (S-T)~Attention on SEED and SEED-IV respectively, indicating comprehensively considering the temporal and spatial characteristics of EEG may boost the emotion recognition results most notably. As for single dimensional attention, the results are a bit of lower than those of combined variants'. Spatial Attention is 0.4\% /0.94\% higher than that of Temporal Attention, implying the spatial dimension may have more emotion-related message than the temporal dimension.  

\section{Conclusion}
In this paper, we propose a new EEG emotion recognition framework based on self-attention, which is built exclusively on self-attention. Our approach considers the relationship between emotion and brain regions, time series change as well as the intrinsic spatiotemporal characteristics of EEG signals. The results of our methods show that the attention mechanism can boost the performance of emotion recognition evidently. Furthermore, the simultaneous spatio-temporal attention gets the best results among the four designed structures, the result is also better than most state of the art methods, indicating that considering the spatio-temporal feature jointly and simultaneously
is more in line with the transmission law of EEG signals in the human brain.

\bibliographystyle{IEEE}
\bibliography{strings,refs}

\end{document}